\documentclass[11pt,a4paper]{article}
\usepackage{times,latexsym}
\usepackage{url}
\usepackage{pbox}
\usepackage[T1]{fontenc}
\usepackage{microtype}
\usepackage{graphicx}
\usepackage{subfigure}
\usepackage{booktabs} % for professional tables
\usepackage{xspace}
\usepackage{ctable}
\usepackage{amssymb}
\usepackage{amsmath}
\usepackage{color,soul}
\usepackage{amsthm}
\usepackage{multirow}
\usepackage{makecell}
\usepackage{enumitem}
\usepackage{mathtools}
\usepackage{calc}
\usepackage{hyperref}
\usepackage{scalerel,stackengine}
\stackMath
\newcommand\reallywidehat[1]{%
\savestack{\tmpbox}{\stretchto{%
  \scaleto{%
    \scalerel*[\widthof{\ensuremath{#1}}]{\kern-.6pt\bigwedge\kern-.6pt}%
    {\rule[-\textheight/2]{1ex}{\textheight}}%WIDTH-LIMITED BIG WEDGE
  }{\textheight}% 
}{0.5ex}}%
\stackon[1pt]{#1}{\tmpbox}%
}

\DeclarePairedDelimiter\norm{\lVert}{\rVert}%

\usepackage[acceptedWithA]{tacl2018v2}

\usepackage{xspace,mfirstuc,tabulary}

\newif\iftaclinstructions
\taclinstructionsfalse % AUTHORS: do NOT set this to true
\iftaclinstructions
\renewcommand{\confidential}{}
\renewcommand{\anonsubtext}{(No author info supplied here, for consistency with
TACL-submission anonymization requirements)}
\newcommand{\instr}
\fi

\iftaclpubformat % this "if" is set by the choice of options

\else

\fi

\newcommand{\hvec}[1]{| {#1} \rangle}
\newcommand{\hcovec}[1]{\langle {#1} |}
\newcommand{\hdot}[2]{\langle {#1} | {#2} \rangle}
\newcommand{\vocab}{\mathcal{V}}
\newcommand{\corpus}{\mathcal{D}}
\newcommand{\hilby}{Hilbert-MLE\xspace}
\newcommand{\hder}{\frac{\partial f_{ij}}{\partial\psi_{ij}}\xspace}
\newcommand{\pmi}{\mathrm{PMI}\xspace}

\title{Deconstructing and reconstructing word embedding algorithms}
\author{
Edward Newell\thanks{These authors contributed equally.}
\qquad
Kian Kenyon-Dean\footnotemark[1]
\qquad
Jackie Chi Kit Cheung\\
Mila - Qu{\'e}bec AI Institute, McGill University\\
Montreal, Canada\\
\texttt{\{edward.newell,kiankd\}@gmail.com} \quad \texttt{jcheung@cs.mcgill.ca}
}
\date{}

\begin{document}

\maketitle
\begin{abstract}
Uncontextualized word embeddings are reliable feature representations of words used to obtain high quality results for various NLP applications. Given the historical success of word embeddings in NLP, we propose a retrospective on some of the most well-known word embedding algorithms. In this work, we deconstruct \textit{Word2vec}, \textit{GloVe}, and others, into a common form, unveiling some of the necessary and sufficient conditions required for making performant word embeddings. 
We find that each algorithm: (1) fits vector-covector dot products to approximate pointwise mutual information (PMI); and, (2) modulates the loss gradient to balance weak and strong signals. 
% We show that training distinct vector and covector representations is a necessary condition for training high-quality word embeddings, suggesting that deeper architectures might benefit from incorporating such a dual representation. 
%
We demonstrate that these two algorithmic features are sufficient conditions to construct a novel word embedding algorithm, \hilby. We find that its embeddings obtain equivalent or better performance against other algorithms across 17 intrinsic and extrinsic datasets.
\end{abstract}

\section{Introduction}
Word embeddings have been established as standard feature representations for words in most contemporary NLP tasks \cite{kim2014convolutional,huang2015bidirectional,goldberg2016primer}. Their incorporation into larger models -- from CNNs and LSTMs for sentiment analysis \cite{zhang2018deep}, to sequence-to-sequence models for machine translation \cite{qi2018and}, to the input layer of deep contextualized embedders \cite{peters2018deep} -- enables high quality performance across a wide variety of problems.

Being the building blocks for many modern NLP applications, we argue that it is worthwhile to subject word embedding algorithms to close theoretical inspection. This work can be considered a retrospective analysis of the ground-breaking word embedding algorithms of the past, which simultaneously offers theoretical insights for how future, deeper models can be developed and understood. Indeed, analogous to a watchmaker who curiously scrutinizes the mechanical components comprising her watches' oscillators, so too do we aim to uncover what makes word embeddings ``tick''. 

It is well-known that word embedding algorithms train two sets of embeddings: the \textit{vectors} (``input'' vectors) and the \textit{covectors} (``output'', or, ``context'' vectors). However, the covectors tend to be regarded as an afterthought when used by NLP practitioners, either being thrown away \cite{mikolov2013distributed}, or averaged into the vectors \cite{pennington2014glove,levy2015improving}. 

Nonetheless, recent work has found that separately incorporating pretrained covectors into downstream models can improve performance in specific tasks. This includes lexical substitution \cite{melamud2015simple,roller2016pic}, information retrieval \cite{nalisnick2016improving}, state-of-the-art metaphor detection \cite{mao2018word} and generation \cite{yu2019avoid}, and more \cite{press2017using,asr2018querying,deugirmenci2019waste}. 
In this work, we contribute an engaged theoretical treatment of covectors, and later elucidate the different relationships learned separately by vectors and covectors (\S\ref{sec:vecscovecs}).
% providing examples that elucidate the difference between the 

Training these vectors and covectors can be done by a variety of high-performing algorithms: the sampling-based shallow neural network of SGNS \cite{mikolov2013distributed}, GloVe's weighted least squares over global corpus statistics \cite{pennington2014glove}, and matrix factorization methods \cite{levy2014neural,levy2015improving,shazeer2016swivel}.  In this work, we propose a framework for understanding these algorithms from a common vantage point.  We deconstruct each algorithm into its constituent parts, and find that, despite their many different hyperparameters, the algorithms collectively intersect upon the following two key design features:
\begin{enumerate}
    \item vector-covector dot products are learned to approximate pointwise mutual information (PMI) statistics in the corpus; and, 
    \item modulation of the loss gradient, directly or indirectly, to balance weak and strong signals arising from the highly imbalanced distribution of corpus statistics.
\end{enumerate}
Finding these commonalities across algorithms, we beg the question of whether or not these features are sufficient for \textit{reconstructing} a new word embedding algorithm. Indeed, we derive and implement a novel embedding algorithm, \hilby\footnote{The name is inspired by the intuitions behind this work concerning Hilbert spaces, and the maximum likelihood estimate that defines the model's loss function.}, by following these two principles to derive their corresponding global matrix factorization loss function based on the maximum likelihood estimate of the multinomial distribution of corpus statistics.

\begin{figure}[t]
    \centering
    \includegraphics[width=\columnwidth]{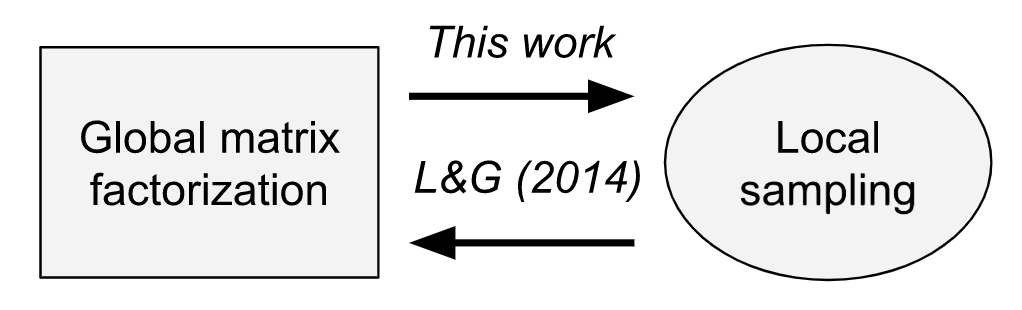}
    \caption{Abstract depiction of the direction of algorithmic derivation presented in this work, in contrast to \citet{levy2014neural}.}
    \label{fig:algs}
\end{figure}

However, due to the infeasibility of matrix factorization objectives for large vocabulary sizes, we further derive a local sampling-based formulation by algebraically deconstructing \hilby's global objective function. As we abstractly depict in Figure~\ref{fig:algs}, this derivation can be seen as a mirrored derivation of that which is presented by \citet{levy2014neural}, who derived the global matrix factorization for SGNS from the original local sampling formulation \cite{mikolov2013distributed}.

We find that \hilby produces word embeddings that earn equivalent or better performance against SGNS and GloVe across 17 intrinsic and extrinsic datasets, therefore demonstrating the sufficiency of the two principles for designing a word embedding algorithm.

To summarize, this work offers the following contributions:
\begin{itemize}
    \item Theoretical deconstruction of the existing dominant word embedding algorithms toward two common features (\S\ref{sec:deconstructing});
    \item Theoretical reconstruction of a novel embedding algorithm, \hilby, based solely on these two features (\S\ref{sec:simple}) which algorithmically derives a sampling-based implementation in a novel manner (\S\ref{sec:reconstructing-solving});
    \item Empirically demonstrating the sufficiency of the two common principles for designing a word embedding algorithm (\S\ref{sec:experiments}).
\end{itemize}

\section{Fundamental concepts}
In this section, we introduce notation and concepts that we will draw upon throughout this paper.  This includes formally defining embeddings, their vectors and covectors, and pointwise mutual information (PMI).

\paragraph{Embedding.}
In general topology, an embedding is understood as an injective ``structure preserving map'', $f: X \rightarrow Y$, between two mathematical structures $X$ and $Y$.  A word embedding algorithm ($f$) learns an inner-product space ($Y$) to preserve a linguistic structure within a reference corpus of text, $\corpus$ ($X$), based on the words in a vocabulary, $\mathcal{V}$.  The structure in $\corpus$ is analyzed in terms of the relationships between words induced by their appearances in the corpus.
In such an analysis, each word figures dually: (1) as a focal element inducing a local context; and (2) as elements of the local contexts induced by focal elements.
To make these dual roles explicit, we distinguish two copies of the vocabulary: the \textit{focal words} $\vocab_T$ (or, \textit{terms}), and the \textit{context words} $\vocab_C$.

An embedding consists of two maps:
\begin{align*}\begin{split}
\vocab_C &\longrightarrow \mathbb{R}^{1 \times d}
\qquad
\vocab_T \longrightarrow \mathbb{R}^{d \times 1}
\\
i &\longmapsto \hcovec{i} \quad\;\:\:\,\qquad j\longmapsto \hvec{j}.
\end{split}\end{align*}
We use Dirac notation to distinguish \textit{vectors} $\hvec{j}$, associated to focal words, from \textit{covectors} $\hcovec{i}$, associated to context words.
In matrix notation, $\hvec{j}$ corresponds to a column vector and $\hcovec{i}$ to a row vector.  Their inner product is $\hdot{i}{j}$; this inner product completely characterizes the learned vector space. We will later demonstrate that many word embedding algorithms, intentionally or not, learn a vector space where the inner product between a focal word $j$ and context word $i$ aims to approximate their PMI in the reference corpus: $\hdot{i}{j} \approx \pmi(i,j)$.

\paragraph{Pointwise mutual information (PMI).}
PMI is a commonly used measure of association in computational linguistics, and has been shown to be consistent and reliable for many tasks \cite{terra2003frequency}. It measures the deviation of the cooccurrence probability between two words $i$ and $j$ from the product of their marginal probabilities:
\begin{equation} \label{eq:pmi}
\begin{split}
    \pmi(i,j) := \ln \frac{p_{ij}}{p_i p_j}
    = \ln \frac{N N_{ij}}{N_i N_j},
\end{split}
\end{equation}
where $p_{ij}$ is the probability of word $i$ and word $j$ cooccurring (for some notion of cooccurrence), and where $p_i$ and $p_j$ are marginal probabilities of words $i$ and $j$ occurring.  The empirical PMI can be found by  replacing probabilities with corpus statistics. Words are typically considered to cooccur if they are separated by no more than $w$ words; $N_{ij}$ is the number of counted cooccurrences between a \textit{context} $i$ and a \textit{term} $j$; $N_i$, $N_j$, and $N$ are computed by marginalizing over the $N_{ij}$ statistics.

\section{Word embedding algorithms} \label{sec:low-rank}
We will now introduce the  \textit{low rank embedder} framework for deconstructing word embedding algorithms, inspired by the theory of generalized low rank models \cite{udell2016generalized}. We unify several word embedding algorithms by observing them all from the common vantage point of their \textit{global loss function}. Note that this framework is only used for theoretical analysis, not practical implementation.

The global loss function for a \textit{low rank embedder} takes the form:
\begin{align}
\mathcal{L} &= \sum_{\mathclap{
    (i,j) \in 
    \vocab_C\!\times\!\vocab_T
}} f_{ij}\Big(
   \;\psi(\hcovec{i}, \hvec{j}),
   \;\;\phi(i,j)\;
\Big),
\label{eq:lre-loss}
\end{align}
and satisfies
\begin{align}
\frac{\partial f_{ij}}{\partial \psi_{ij}} =  0
\quad\text{at}\quad
\psi_{ij} = \phi_{ij},
\label{eq:deriv-constraint}
\end{align}
where $\psi(\hcovec{i}, \hvec{j})$ is a kernel function, and $\phi(i,j)$ is some scalar function (such as a measure of association based on how $i$ and $j$ appear in the corpus); $\psi_{ij}$ and $\phi_{ij}$ are abbreviations for the same.

The design variable $\phi_{ij}$ is some function of \textit{corpus statistics}, and its purpose is to  quantitatively measure some relationship between word $i$ and $j$.
In apposition, the design variable $\psi_{ij}$ is a function of \textit{model parameters}, and its purpose is to learn a succinct approximation:
\begin{align*}
\psi_{ij} \approx \phi_{ij},
\end{align*}
and so represent the relationship measured by  $\phi_{ij}$.  
Intuitively, we can think of a low rank embedder as trying to directly fit a kernel function of model parameters $\psi_{ij}$ to some (statistical) relationship between words $\phi_{ij}$ of our choosing.  For example, SGNS takes  $\phi_{ij} := \pmi(i,j) - \ln k$ and $\psi_{ij} := \hdot{i}{j}$, and then learns parameter values that approximate $\hdot{i}{j}\approx \pmi(i,j) - \ln k$.

Though the specific choice of $\phi_{ij}$ varies slightly, existing low rank embedders generally base $\phi_{ij}$ on cooccurrence
of words within a linear window $w$ words wide.
But it is worth pointing out that $\phi_{ij}$ can in principle be any pairwise relationship encoded as a scalar function of corpus statistics.

As for the kernel function $\psi_{ij}$, one simple choice is to take $\psi_{ij} = \hdot{i}{j}$. But the framework allows any function that is symmetric and positive definite. This allows the framework to include the use of bias parameters (e.g. in GloVe) and subword parameterization (e.g. in FastText).

To understand the range of models encompassed, it is helpful to see how the framework relates (but is not limited) to matrix factorization.  
We can think of $\phi_{ij}$ and $\psi_{ij}$ as providing the entries of two matrices: 
\begin{align*}
\mathbf{M} := \left[\phi_{ij}\right]_{ij}
\quad
\hat{\mathbf{M}} := \left[\psi_{ij}\right]_{ij}.
\end{align*}
For models that take $\psi_{ij} = \hdot{i}{j}$, we can write $\hat{\mathbf{M}} = \mathbf{WV}$, where $\mathbf{W}$ is defined as having row $i$ equal to $\hcovec{i}$, and $\mathbf{V}$ as having column $j$ equal to $\hvec{j}$.
Then, the loss function can be rewritten as:
\begin{align*}
\mathcal{L} 
%&= \sum_{\mathclap{
%    (i,j) \in 
%    \vocab_C\!\times\!\vocab_T
%}} f_{ij}\Big(
%   \hat{\mathbf{M}}_{ij},
%   \mathbf{M}_{ij}
%\Big),
%\\
&= \sum_{\mathclap{
    (i,j) \in 
    \vocab_C\!\times\!\vocab_T
}} f_{ij}\Big(
   (\mathbf{WV})_{ij},\;
   \mathbf{M}_{ij}
\Big).
\end{align*}
This loss function can be interpreted as matrix reconstruction error, because the constraint in Eq.~\ref{eq:deriv-constraint} means that the gradient goes to zero as $\mathbf{WV} \approx \mathbf{M}$.

% It should be noted that the framework admits algorithms that one might not normally consider as matrix factorizations per se.  SGNS is an example of an algorithm that \textit{implicitly} factorizes a matrix without explicitly representing it in memory, and we discuss links between implicit and explicit matrix factorization later (\S\ref{sec:densesparse}).

Selecting a particular low rank embedder instance requires key design choices to be made:
we must chose the embedding dimension $d$, the form of the loss terms $f_{ij}$, the kernel function $\psi_{ij}$, and the association function $\phi_{ij}$.  
Only the \textit{gradient} of $f_{ij}$ actually affects the algorithm.  The derivative of $f_{ij}$ with respect to $\psi_{ij}$, which we call the \textit{characteristic gradient}, helps compare models because it exhibits the action of the gradient yet is symmetric in the parameters.
Thus, we address a specific embedder by the tuple $\left(d, \hder, \psi_{ij}, \phi_{ij}, \right)$.

In the following subsections, we present the derivations of $\hder$, $\psi_{ij}$, and $\phi_{ij}$ for each algorithm. We later (\S\ref{sec:deconstructing}) compare the algorithms and summarize their derivations in Table~\ref{tab:embedder-form-comparison}.

% We derive the forms for SGNS and LDS in \S\ref{sec:additional-derivations}.
% The forms for GloVe  follow directly from its loss function (see \S\ref{sec:mfglv}).
% The entries for Swivel can be trivially derived form the original formulation \cite{shazeer2016swivel}.  The forms for \hilby are derived in \S\ref{sec:simple}.
% We also show that FastText fits into the framework in \S\ref{sec:additional-derivations}, but omit it from the table because it is very similar to SGNS.   

\subsection{SGNS as a low rank embedder} \label{sec:sgns}
\citet{levy2014neural} provided the important result that \textit{skip gram with negative sampling} (SGNS) \cite{mikolov2013distributed} was implicitly factorizing the $\pmi - \ln k$ matrix. However, \citeauthor{levy2014neural} did not derive the loss function needed to explicitly pose SGNS as matrix factorization, and required additional assumptions for their derivation to hold. Moreover, empirically, they used SVD on the related (but different) \textit{positive}-PMI matrix, and this did not reproduce the results of SGNS.
In other work, \citet{li2015word}, provided an explicit MF formulation of SGNS from a ``representation learning'' perspective.  This derivation diverges from \citeauthor{levy2014neural}'s result, and masks the connection between SGNS and other low rank embedders. In this work, we derive the complete global loss function for SGNS, free of additional assumptions.

The loss function of SGNS is as follows:
\begin{equation*}
\mathcal{L} = - \sum_{\mathclap{(i,j) \in D_2}}
\Big\{
\ln \sigma \hdot{i}{j} + \sum_{\ell=1}^k \mathbb{E} \Big[ 
\ln (1 - \sigma \hdot{i'_\ell}{j})
\Big]
\Big\},
\end{equation*}
where $\sigma$ is the logistic sigmoid function, $D_2$ is a \textit{list} containing each cooccurrence of a context-word $i$ with a focal-word $j$ in the corpus, and the expectation is taken by drawing $i'_\ell$ from the (smoothed) unigram distribution to generate $k$ ``negative samples'' for a given focal-word.   \cite{mikolov2013distributed}.
% ($D_2$ is a list in the sense that given pairs $(i,j)$ are repeated as many times as they cooccur in the reference corpus.)

We rewrite this by counting the number of times each pair occurs in the corpus, $N_{ij}$, and the number of times each pair is drawn as a negative sample, $N_{ij}^-$, while indexing the sum over the set $\vocab_C \!\times\!\vocab_T$:
\begin{equation*} 
\mathcal{L} = - \sum_{\mathclap{
    (i,j) \in 
    \vocab_C\!\times\!\vocab_T
}} 
\Big\{
N_{ij}\ln \sigma \hdot{i}{j} +  
N_{ij}^- \ln( 1- \sigma \hdot{i}{j})
\Big\}.
\end{equation*}
%This can be viewed as a loss-function computed over the elements of the matrix product $\textbf{VW}$ with elements $(\mathbf{VW})_{ij} = \hdot{i}{j}$.  
%The target matrix being factorized can be seen by examining the partial derivative of the loss function with respect to this inner product, $\hder$:
%

We can now observe that the global loss is almost in the required form for a low rank embedder (Eq.~\ref{eq:lre-loss}), and that the appropriate setting for the model approximation function is $\psi_{ij} = \hdot{i}{j}$.
The characteristic gradient is derived as, using the identity $a \equiv (a+b)\sigma(\ln \frac{a}{b})$:
\begin{align*}
\hder &= \frac{\partial\mathcal{L}}{\partial\hdot{i}{j}}
   = N_{ij}^- \sigma \hdot{i}{j} - N_{ij} (1 - \sigma \hdot{i}{j}) \\
   &= (N_{ij} + N_{ij}^-) \bigg[ \sigma \big( \hdot{i}{j} \big)
- \sigma \big( \ln\frac{N_{ij}}{N_{ij}^-} \big) \bigg]. \label{eq:sgns-M}
\end{align*}
This provides that the association function for SGNS is $\phi_{ij} = \ln(N_{ij}/N_{ij}^-)$, since the derivative will be equal to zero at that point (Eq.~\ref{eq:deriv-constraint}). However, recall that negative samples are drawn according to the unigram distribution (or a smoothed variant \cite{levy2015improving}). This means that $N_{ij}^- = kN_iN_j/N$. Therefore, in agreement with \citet{levy2014neural}, we find that: 
\begin{align}
    \phi_{ij} = \ln \frac{N_{ij} N}{N_i N_j k} = \pmi(i,j) - \ln k.
\end{align}

\subsection{GloVe as a low rank embedder} \label{sec:glove}
GloVe (global vectors) was proposed as a method that strikes a halfway point between local sampling and global matrix factorization, taking the best parts from both solution methods \cite{pennington2014glove}. Its efficiency came from the fact that it only performs a partial factorization of the $\ln N_{ij}$ matrix, only considering samples where $N_{ij} > 0$. We will demonstrate that GloVe is not so different from SGNS, and that it too implicitly factorizes the $\pmi$ matrix.

GloVe's loss function is defined as follows:
\begin{align}
\begin{split}
\mathcal{L} = 
   \sum_{ij} h(N_{ij}) \Big(\hdot{i}{j} + b_i + b_j - \ln N_{ij} \Big)^2 \\
h(N_{ij}) = 
   \min\left(1, \left(\frac{N_{ij}}{N_\mathrm{max}}\right)^{\alpha}\right),\qquad
\end{split}
\end{align}
where $b_i$ and $b_j$ are learned bias parameters; $N_\mathrm{max}$ and $\alpha$ are empirically tuned hyperparameters for the weighting function $h(N_{ij})$, which has $h(N_{ij}) = 0$ when $N_{ij} = 0$.

GloVe can be cast as a low rank embedder by using the model approximation function as a kernel with bias parameters, and setting the association measure to simply be the objective of the loss function:
\begin{align*}
\psi_{ij} = \big[
   \,\hcovec{i}_1 \, \cdots \, \hcovec{i}_d \: b_i \; 1\,
\big]
&\cdot
\big[
   \,\hvec{j}_1 \cdots \hvec{j}_d \:\: 1 \:\: b_j\:
\big]^\intercal,
\\
\text{and}\quad
\phi_{ij}&=\ln N_{ij}.
\end{align*}

\begin{figure}[t]
    \centering
    \includegraphics[width=\columnwidth]{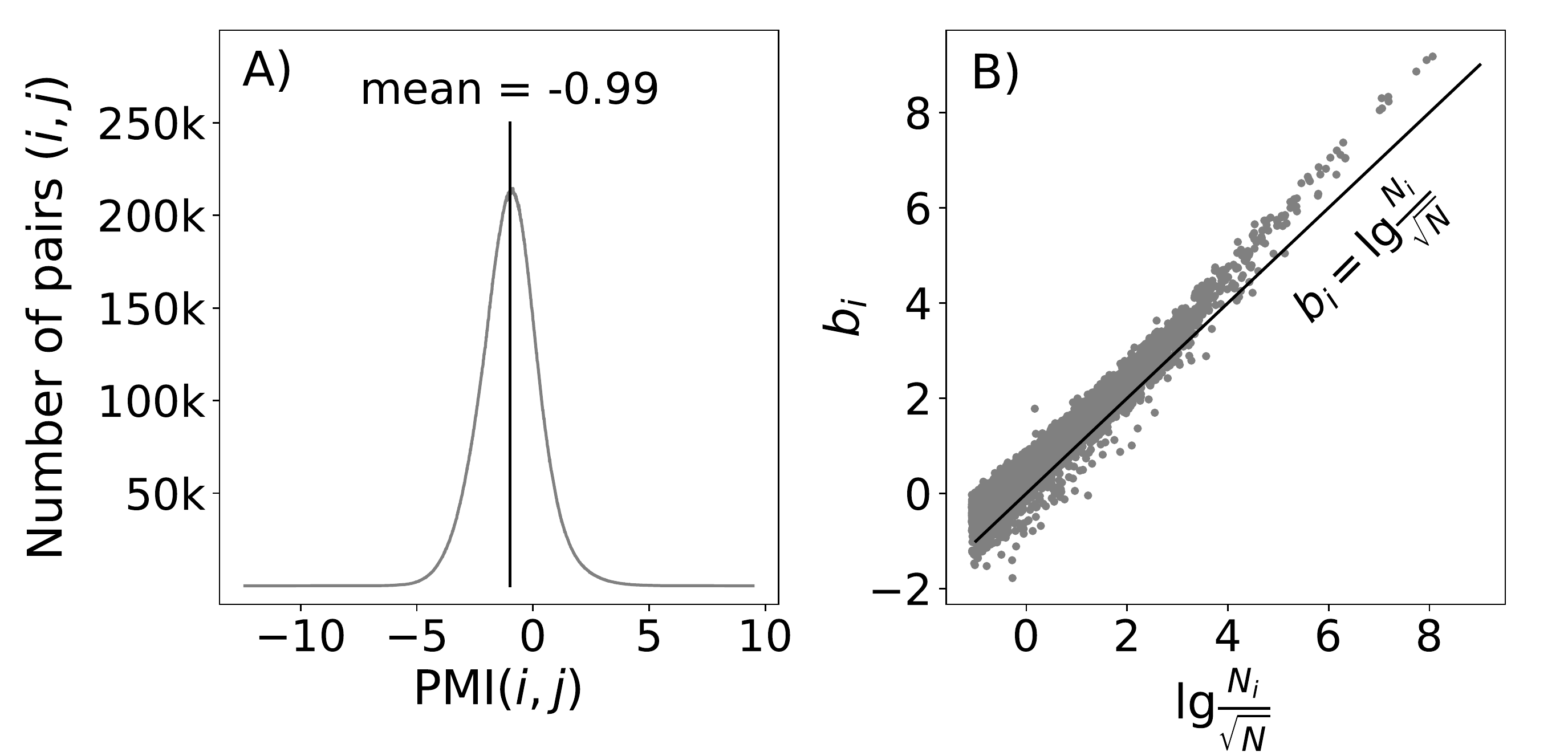}

    \caption{\textbf{A}) Histogram of $\pmi(i,j)$ values, for all pairs $(i,j)$ with $N_{ij}\!>\!0$, from the corpus described in \S\ref{sec:experiments}. \textbf{B}) Scatter plot of GloVe's learned biases after $10$ epochs, using default hyperparameter settings.}
    \label{fig:glove-is-pmi}
\end{figure}

Let us observe the optimal solution to the loss function, when $\hder = 0$:
\begin{align*}
\hder &= 2h(N_{ij})\Big[\hdot{i}{j} + b_i + b_j - \ln N_{ij}\Big] = 0\\
&\implies
    \hdot{i}{j} + b_i + b_j = \ln N_{ij}.
\end{align*}
Multiplying the log operand by $1$:
\begin{align}
    \hdot{i}{j} + b_i &+ b_j = \ln \left(\frac{N_iN_j}{N}\frac{N}{N_iN_j}N_{ij}\right) \\
    &= \ln \frac{N_i}{\sqrt{N}} + \ln\frac{N_j}{\sqrt{N}} + \mathrm{PMI}(i,j) .
\label{eq:glove-is-pmi}
\end{align}
On the right side, we have two terms that depend respectively only on $i$ and $j$, which are candidates for the bias terms.  Based on this equation alone, we cannot draw any conclusions.  
However, empirically the bias terms are in fact very near $\frac{N_i}{\sqrt{N}}$ and $\frac{N_j}{\sqrt{N}}$, and
PMI is nearly centered at zero, as can be seen in 
Fig.~\ref{fig:glove-is-pmi}.
This means that 
Eq.~\ref{eq:glove-is-pmi} provides $\hdot{i}{j} \approx \pmi(i,j)$.

% The fact that the distribution of PMI is approximately centered can be understood intuitively.  Fix a focal word $j$, and consider $\pmi(i,j)$ for various context words $i$.  Any $i$ cooccurring more often with $j$ than would occur randomly assuming independence---hence having $\pmi(i,j)>0$---does so at the expense of other context word(s) $i'$ which must have $\pmi(i',j)<0$.  

Analyzing the optimum of GloVe's loss function yields important insights.  First, GloVe can be added to the list of low rank embedders that learn a bilinear parameterization of PMI.  Second, we can see why such a parameterization is advantageous.  Generally, it helps to standardize features of low rank models \cite{udell2016generalized}, and this is essentially what transforming cooccurrence counts into PMI achieves.  Thus, PMI can be viewed as a parameterization trick, providing an approximately normal target association to be modelled.

\subsection{Other algorithms as low rank embedders} \label{sec:otheralgs}
We now present additional algorithms that can be cast as low rank embedders: LDS \cite{arora2016latent} and FastText \cite{joulin2017bag}. The derivations for SVD \cite{levy2014neural,levy2015improving} and Swivel \cite{shazeer2016swivel} as low rank embedders are trivial, as both are already posed as matrix factorizations of PMI statistics.

\paragraph{LDS.} \citet{arora2016latent} introduced an embedding perspective based on generative modelling with random walks through a latent discourse space (LDS). While their only experiments were on analogy completion tasks (which do not correlate well with downstream performance \cite{linzen2016issues,faruqui2016problems,rogers2017too}) LDS provided a theoretical basis for the surprisingly well-performing \textit{SIF document embedding} algorithm soon afterwards \cite{arora2017simple}. We now demonstrate that LDS is also a low-rank embedder.
% ote, however, that \citeauthor{arora2016latent}'s only experiments were on analogy tasks, which do not correlate well with performance on practical NLP problems \cite{linzen2016issues,faruqui2016problems,rogers2017too}, and despite including additional hyperparameters, their model showed no improvement over other algorithms. 

The low rank learning objective for LDS follows directly from \textbf{Corollary 2.3}, in \citet{arora2016latent}:
\begin{equation*}
\begin{split}
    \pmi(i,j) &= \frac{\hdot{i}{j}}{d} + \gamma + O(\epsilon).
\end{split}
\end{equation*}
$\hder$ can be found by straightforward differentiation of LDS's loss function: 
\begin{equation*}
\mathcal{L} = \sum_{ij} h(N_{ij})\big[\ln N_{ij} - 
    \norm*{\hcovec{i}+\hvec{j}^\intercal}^2 -C \big]^2,
\end{equation*}
where $h(N_{ij})$ is as defined by GloVe.
The quadratic term is a valid kernel function because:
\begin{align*}
\hder = \norm*{\hcovec{i} + \hvec{j}^\intercal}^2 = 
\langle\tilde{i}|\tilde{j}\rangle,
\end{align*}
where
\begin{align*}
\tilde{\hcovec{i}} &= \Big[\sqrt{2}\hcovec{i}_1\:\,\cdots\;\sqrt{2}\hcovec{i}_d\;\;\hcovec{i}\hcovec{i}^\intercal\;\;\;\;1\;\;\;\;\;\Big]
,\\
\tilde{\hvec{j}} &= 
\Big[
\sqrt{2}\hvec{j}_1\;\cdots\;
\sqrt{2}\hvec{j}_d\;\;\;\,\;1\;\;\;\;\hvec{j}^\intercal\hvec{j}\;\Big]^\intercal.
\end{align*}

\paragraph{FastText.} Proposed by \citet{joulin2017bag}, FastText's motivation is orthogonal to the present work. It's purpose is to provide subword-based representation of words to improve vocabulary coverage and generalizability of word embeddings. Nonetheless, it can also be understood as a low rank embedder.

In FastText, the vector for each word is taken as the sum of embeddings for its character $n$-grams, $3 \leq n \leq 6$. Then the vector $\hvec{j}$ is given by the feature function $\hvec{j} = \sum_{g \in z(j)} \hvec{g}$, where $\hvec{g}$ is the vector for $n$-gram $g$, and $z(j)$ is the set of $n$-grams in word $j$.  Meanwhile covectors are accorded to words directly, rather than using $n$-gram covector embeddings. This provides $\psi_{ij} = \hdot{i}{j}$, and, by virtue of using the SGNS loss function, $\phi_{ij} = \pmi(i,j) - \ln k$.

% a matrix whose rows are $n$-gram embeddings, and define the $n$-gram selector matrix $\mathbf{G}$:
% \begin{align*}
% \mathbf{G}_{ij} := 
% \begin{cases}
% 1 & \text{if $n$-gram $g_j$ is in word $i$} \\
% 0 & \text{otherwise}.
% \end{cases}
% \end{align*}
% Then, the loss function is calculated elementwise over the matrix product $\mathbf{W}\mathbf{G}\mathbf{V}$.
% FastText is otherwise like SGNS, so we do not  include it in Table~\ref{tab:embedder-form-comparison}.

\setlength{\tabcolsep}{5pt}
\begin{table*}[t]
\centering
\begin{tabular}{c|l|c|c|c}
\textbf{Model} &
\multicolumn{1}{c|}{$\hder$}
& 
$\psi_{ij}$
&
$\phi_{ij}$
&
$\hdot{i}{j} \approx$
\rule{0pt}{0.8em}\rule[-0.8em]{0em}{0em}\\ \Xhline{3\arrayrulewidth}

%%%%%%%%%%%%% SGNS row %%%%%%%%%%%%% 

SGNS 

&

\rule{0pt}{0em}
$(N_{ij} + N_{ij}^-) \cdot \big[ 
\sigma (\psi_{ij})
- \sigma(\phi_{ij})
\big]$

& 

$\hdot{i}{j}$

&

$\lg\frac{N_{ij}}{N_{ij}^-}$

&

$\pmi(i,j) - \ln k$

\rule{0pt}{1.5em}\rule[-1.1em]{0em}{0em}\\ \hline

%%%%%%%%%%%%% Glove row %%%%%%%%%%%%% 
GloVe

&

\rule{1.7em}{0em}
$2 h(N_{ij}) \cdot \big[\psi_{ij} - \phi_{ij} \big]\quad\;$

&

$\hdot{i}{j} + b_i + b_j$

& 

$\lg N_{ij}$

&

$\pmi(i,j)$

\rule{0pt}{1.5em}\rule[-1.0em]{0em}{0em}\\ \hline

%%%%%%%%%%%%% LDS row %%%%%%%%%%%%% 
LDS
& 
\rule{1.72em}{0em}
$4 h(N_{ij})
\cdot
\Big[
\psi_{ij}
-\phi_{ij} + C
\Big]$
&
$\norm{\hcovec{i}+\hvec{j}^\intercal}^2$
&
$\lg N_{ij}$
&
$d \pmi(i,j) - d\gamma$
\rule{0pt}{1.5em}\rule[-1.0em]{0em}{0em}\\ \hline

%%%%%%%%%%%%% Swivel row %%%%%%%%%%%%% 

\multirow{2}{*}{Swivel\rule{0em}{1.8em}}
& 
\rule{2.52em}{0em}
$\sqrt{N_{ij}} \cdot \Big[ \psi_{ij} - \phi_{ij} \Big]$
&
\multirow{2}{*}{
$\hdot{i}{j}$
\rule{0em}{1.8em}}
&
$\pmi(i,j)$
&
$\pmi(i,j)$

\rule{0pt}{1.5em}\rule[-1.0em]{0em}{0em}\\ 
& 
\rule{4.48em}{0em}
$1\cdot \sigma \Big(\psi_{ij} - \phi_{ij} \Big)$
&
&
$\pmi^*(i,j)$
&
$\pmi^*(i,j)$
\rule{0pt}{0em}\rule[-.8em]{0em}{0em}\\[2mm] \Xhline{3\arrayrulewidth}

%%%%%%%%%%%%% Hilby row %%%%%%%%%%%%% 

%\pbox{2cm}{\centering Hilbert-MLE}
\hilby

& 
\rule{1.69em}{0em}
$\big(p_i p_j\big)^\frac{1}{\tau} \cdot \Big[
e^{\psi_{ij}}
- e^{\phi_{ij}}
\Big]$
&
$\hdot{i}{j}$
&
$\pmi(i,j)$
&
$\pmi(i,j)$

\rule{0pt}{1.5em}\rule[-0em]{0em}{0em}\\ 

    \end{tabular}
    \caption{Comparison of low rank embedders.  Final column shows the value of $\hdot{i}{j}$ at $\hder=0$.
    GloVe and LDS set $f_{ij}=0$ when $N_{ij}=0$.  Swivel takes one form when  $N_{ij}>0$ (first row) and another when $N_{ij}=0$ (second row).
    $N_{ij}^-$ is the number of negative samples.
    For other symbols see: SGNS \cite{mikolov2013distributed}, GloVe \cite{pennington2014glove}, LDS \cite{arora2016latent}, Swivel \cite{shazeer2016swivel}.
    }
    \label{tab:embedder-form-comparison}
\end{table*}
\section{Deconstructing the algorithms} \label{sec:deconstructing}
Table \ref{tab:embedder-form-comparison} presents a summary of our derivations of existing algorithms as low rank embedders. 

We observe several common features between each of the algorithms. In each case, $\hder$ takes the form $(\mathrm{multiplier})\cdot(\mathrm{difference})$. 
The multiplier is always a ``tempered'' version of $N_{ij}$ (or $N_iN_j$), by which we mean that it increases sublinearly with $N_{ij}$ (or $N_iN_j$)\footnote{In SGNS, $N_{ij}^- \propto N_iN_j$; $N_{ij}$ and $N_{ij}^-$ are tempered by undersampling and unigram smoothing.}.

Furthermore, for each algorithm, $\phi_{ij}$ is equal to PMI or a scaled log of $N_{ij}$. Yet, the choice of $\psi_{ij}$ in combination with $\phi_{ij}$ provides that every model is optimized when $\hdot{i}{j}$ tends toward $\pmi(i,j)$ (with or without a constant shift or scaling). We have already seen that the optimum for SGNS is equivalent to the shifted PMI (\S\ref{sec:sgns}). For GloVe, we theoretically and empirically showed that incorporation of the bias terms captures the unigram counts needed for PMI (\S\ref{sec:glove}). We observe this property similarly with regards to LDS's incorporation of the L2 norm into its learning objective, where we suspect that the unigram probability is implicitly captured in the norms of the respective vectors and covectors (\S\ref{sec:otheralgs}).

Therefore, we observe that these embedders converge on two key points: (1) an optimum in which model parameters are bilinearly related to PMI, and (2) the weighting of $\hder$ by some tempered form of $N_{ij}$ (or $N_iN_j$).  In the next section, we introduce \hilby, which is derived based on the shared principles observed between the algorithms in Table~\ref{tab:embedder-form-comparison}.

\section{Reconstructing an algorithm}
\label{sec:simple}
If the two basic principles that we have identified are sufficient, then the simplest low rank embedder should be one that derives from them without any other assumptions.  

We begin with principle (1), which prescribes a bilinear parameterization of PMI. 
The definition of PMI (Eq.~\ref{eq:pmi}) provides a log-bilinear parameterization of cooccurrence probability, $\hat{p}_{ij}$, if we presuppose that the aim of our model is to approximate the PMI with vector-covector dot products: 
\begin{align}
\begin{split}
\hdot{i}{j} &\approx \pmi(i,j) = \ln \frac{p_{ij}}{p_i p_j} \\
\implies \quad \hat{p}_{ij} &= p_ip_je^{\hdot{i}{j}}.
\label{eq:bilinear-pmi}
\end{split}
\end{align}
In the expression above, $\hat{p}_{ij}$ represents the model's estimate of the cooccurrence probability, provided by the parameterization which includes the unigram probabilities $p_i$ and $p_j$.

Accordingly, given the matrix of covectors $\mathbf{W}$ and vectors $\mathbf{V}$, the likelihood of the observed cooccurrence statistics, 
$\mathcal{D}=\{N_{ij}\}_{ij}$,
is distributed like the multinomial,  $\mathrm{Mult}(\{\hat{p}_{ij}\}_{ij},N)$:
\begin{align*}
\textrm{Pr}(\corpus | \mathbf{V}, \mathbf{W})
 = N!\prod_{ij} \frac{\hat{p}_{ij}^{N_{ij}}}{N_{ij}!},
\end{align*}
where $\hat{p}_{ij}$ depends on $\mathbf{V}$ and $\mathbf{W}$ (whose rows and columns are respectively $\hcovec{i}$ and $\hvec{j}$) through Eq.~\ref{eq:bilinear-pmi}. 
Taking the negative log likelihood as the loss:
\begin{align}
\mathcal{L} = -\sum_{ij} N_{ij} \ln \hat{p}_{ij},
\end{align}
where we have dropped constant terms that do not affect the gradient.  

The unitarity axiom of probability requires that $\sum_{ij}\hat{p}_{ij} = 1$.  Including this constraint with a Lagrange multiplier, we obtain:
\begin{align}
\mathcal{L} = \Big(\!-\!\sum_{ij} 
      N_{ij} \ln \hat{p}_{ij} \Big)
   + \lambda\Big(
      1-\sum_{ij}\hat{p}_{ij}\Big).
\end{align}
At the feasible optimum, the original loss and constraint gradients should balance:
\begin{align}
\frac{\partial\mathcal{L}}{\partial \hat{p}_{ij}} = 
-\frac{N_{ij}}{\hat{p}_{ij}} - \lambda = 0 \\
\implies \lambda \hat{p}_{ij} = -N_{ij}.
\label{eq:lambda}
\end{align}
Eq.~\ref{eq:lambda} represents $|\vocab_C\!\times\!\vocab_T|$ equations, one for each pair $(i,j)$.  Summing these equations together, 
\begin{align}
\lambda \sum_{ij} \hat{p}_{ij} &= -\sum_{ij} N_{ij} \;\;\;
\implies \;\;\; \lambda = -N.
\end{align}
The constrained loss function is therefore,
\begin{align}
\mathcal{L} = \Big(
\!-\!\sum_{ij} N_{ij}\ln \hat{p}_{ij}\Big) - N\Big(1\!-\!\sum_{ij} \hat{p}_{ij}\Big).
\end{align}
Reintroducing the bilinear parameterization (Eq.~\ref{eq:bilinear-pmi}), and dividing through by $N$ to eliminate dependence on corpus size:
\begin{align}
\mathcal{L} = \sum_{ij}\Big( p_ip_je^{\hdot{i}{j}} -
p_{ij} \hdot{i}{j} 
\Big),\quad
\label{eq:final-loss}
\end{align}
where, again, we have dropped constant terms that do not affect the gradient.
Finally by differentiating we obtain the characteristic gradient:
\begin{align}
\begin{split}
\hder = \frac{\partial f_{ij}}{\partial\hdot{i}{j}} &= p_ip_j\Big[
e^{\hdot{i}{j}} -
e^\mathrm{PMI(i,j)}
\Big]\\
&= \hat{p}_{ij} - p_{ij}.
\end{split}
\label{eq:hilby-gradient}
\end{align}
This yields a loss gradient closely resembling other members of the low rank embedders.  Empirically, its performance is on par with the other low rank embedders (see \S\ref{sec:experiments}).

The multiplier, $p_ip_j$, determines how errors in fitting individual $(i,j)$ pairs trade off.  While it appropriately favors fitting statistics with lower standard error, the signal from rarer pairs will be weak for any non-divergent learning rate because $p_ip_j$ spans orders of magnitude.  This slows down training. So, we apply a gradient conditioning measure as is done for the other low rank embedders: we apply a temperature parameter, $\tau$, that reduces differences in magnitude of the multiplier:
\begin{align}
\hder &= (p_ip_j)^{1/\tau}
\Big[
e^{\hdot{i}{j}} -
e^\mathrm{PMI}
\Big]
\label{eq:hilby-gradient-tau}
\end{align}

% Empirically, we found tuning to $\tau=2$ provides good results on word similarity experiments and fast convergence.

\subsection{Solving the objective function} \label{sec:reconstructing-solving}
The objective function presented in Equation~\ref{eq:final-loss} is most straightforwardly solved via dense matrix factorization. This can be done relatively efficiently by using the sharding method presented by \citet{shazeer2016swivel} for matrix factorization. Such a solution is acceptable given a small vocabulary size, but does not scale to large vocabularies, due to the quadratic dependency. GloVe \cite{pennington2014glove} handled this problem by only training on statistics where $N_{ij} > 0$. \citet{levy2014neural,levy2015improving} avoided the quadratic dependency by implementing sparse SVD on the positive-PMI matrix. However, both of these solutions may be missing out on important information that can be gained by ``noticing what's missing'' \cite{shazeer2016swivel}.

Yet, SGNS \cite{mikolov2013distributed} was never confronted with the vocabulary size problem due to the fact that it uses \textit{local sampling} over the corpus. While this yields a linear time complexity on the corpus size, this is generally preferable to a quadratic memory complexity on the vocabulary size. \citet{levy2014neural} derived the global matrix factorization formulation of SGNS by moving in the algorithmic direction of \textit{local to global}. Conversely, we will now move in the direction of \textit{global to local} and derive the local sampling formulation of the \hilby loss function.

\paragraph{Locally sampling \hilby.} Note how if we differentiate the loss function of \hilby (Equation~\ref{eq:final-loss}) relative to an arbitrary model parameter $\theta$, we obtain a difference between two expectations: 
\begin{equation}
\begin{split}
    \frac{\partial\mathcal{L}}{\partial\theta} 
    \!&=\!
    \sum_{ij}p_ip_je^{\hdot{i}{j}}
    \frac{\partial\hdot{i}{j}}{\partial\theta}
    \!-\!
    \sum_{ij} p_{ij}\frac{\partial\hdot{i}{j}}{\partial\theta} \\
    &= \quad\mathop{\mathbb{E}}_{\mathclap{(i,j)\sim \hat{p}_{ij}}}\quad \left[\frac{\partial\hdot{i}{j}}{\partial\theta}\right] \!- \quad\mathop{\mathbb{E}}_{\mathclap{(i,j)\sim p_{ij}}}\quad \left[\frac{\partial\hdot{i}{j}}{\partial\theta}\right].
\end{split}
\end{equation}
%\begin{equation}
%    \mathcal{L} = \mathbb{E}_{(i,j) \sim p_{ij}} \big[\hdot{i}{j}\big] - \mathbb{E}_{(i,j) \sim \hat{p}_{ij}} \big[\hdot{i}{j}\big]
%\end{equation}
In words, the derivative of the loss function is the difference between the expected value of $\frac{\partial\hdot{i}{j}}{\partial\theta}$ when taken under the model distribution on one hand and under the corpus distribution on the other.

This leads to a remarkably simple training algorithm.  Draw a sample of word pairs $(i,j)$ from the corpus (using a local sampling approach as in SGNS), and draw a sample of pairs from the model distribution.  Compute $\mathbb{E}[\hdot{i}{j}]$ for both samples, and take their difference.  The gradient of the result estimates the gradient of $\mathcal{L}$.

In the context of autodifferentiation libraries such as PyTorch \cite{paszke2017automatic} it is adequate to use a simplified loss function $\mathcal{\tilde{L}}$,
\begin{equation}
    \mathcal{\tilde{L}}
    = \quad\mathop{\mathbb{E}}_{\mathclap{(i,j)\sim \hat{p}_{ij}}}\quad \left[\hdot{i}{j}\right] - \quad\mathop{\mathbb{E}}_{\mathclap{(i,j)\sim p_{ij}}}\quad \left[\hdot{i}{j}\right],
\end{equation}
because in the first term, the autodifferential operator will ignore the appearance of model parameters in the distribution according to which the expectation $\mathbb{E}_{(ij)\sim\hat{p}_{ij}}$ is taken, but will recognize model parameters in the expectation's operand $\hdot{i}{j}$.  Thus
$\frac{\partial_{\text{auto}}}{\partial\theta}\mathcal{\tilde{L}} = \frac{\partial}{\partial\theta}\mathcal{L}$.

Like SGNS, this uses positive samples drawn from the corpus, balanced against negative samples.  But unlike SGNS, which draws negative samples according to the (distorted) unigram distribution, here we draw negative samples from model distribution $\hat{p}_{ij}$.  This can be done efficiently using Gibbs sampling, making this a form of contrastive divergence \cite{hinton2002training, carreira2005contrastive}.  To approximately sample a cooccurring pair $(i,j)$ from the model distribution, we start from a corpus-derived pair, and repeatedly perform Gibbs sampling steps: randomly fix either $i$ or $j$ and re-sample the other from the model distribution conditioned on the fixed variable.  E.g. if we fix $i$, then we draw a new $j$ from $j\sim \hat{p}(j|i)= p_j e^{\hdot{i}{j}}$.  Sampling from the conditional distribution can be done in constant time using an adaptive softmax \cite{grave2017efficient}.
In theory, the model distribution is approximated after taking many Gibbs sample steps, but consistent with Hinton's findings for contrastive divergence  \cite{hinton2002training, carreira2005contrastive}, we find that a single Gibbs sampling step supports efficient training.

% \begin{align}
% \mathcal{L} = \quad
% \;\mathop{\mathbb{E}}_{(i,j)\sim p_ip_je^{\hdot{i}{j}}}\{
%     \hdot{i}{j}\}
% \;-\;\;\;     
% \mathop{\mathbb{E}}_{\mathclap{(i,j)\sim p_{ij}}}\{ 
%     \hdot{i}{j} \}.
% \end{align}
% These expectations can be calculated by sampling pairs from the respective distributions: i.e. using ``positive'' samples from the corpus, and ``negative'' samples from the unigram distribution\footnote{Note how the negative sampling term originates from the unitarity constraint.}.
% To emphasize the similarity of this to the sample-based SGNS algorithm (c.f. \S\ref{sec:mfsgns}), we can rewrite Eq.~\ref{eq:final-loss} yet again as:
% %
% \begin{align}
% \mathcal{L} = -\sum_{(i,j) \in D_2} \Big[ \hdot{i}{j} - \mathop{\mathbb{E}}_{i'\sim p_i}\{e^{\hdot{i'}{j}}\}\Big],
% \end{align}
% where $D_2$ is a list of all of the $(i,j)$ pairs, as they occur when reading through the corpus. 

% 5,389,537,643 tokens

\section{Experiments} \label{sec:experiments}
We provide a simple set of experiments comparing the two characteristic models for word embeddings with ours: SGNS and GloVe against \hilby. Our aim in these experiments is simply to verify the sufficiency of the principles we used to derive \hilby (\S\ref{sec:simple}). In other words, we are testing the following \textbf{hypothesis}: \textit{if the principles we have proposed are sufficient for designing a word embedding algorithm, then \hilby should perform equivalently or better than SGNS and GloVe, which were proposed with different motivating principles} \cite{mikolov2013distributed,pennington2014glove}.

In our experiments, we use a matrix factorization implementation of \hilby as the characteristic form of the model. During experimentation, we found that the Gibbs sampling implementation of \hilby (\S\ref{sec:reconstructing-solving}) performed equivalently, as expected.
We present results on word similarity (\S\ref{sec:wordsim}), analogical reasoning (\S\ref{sec:analogy}), text classification (\S\ref{sec:classification}), and sequence labelling (\S\ref{sec:seqlab}).

Our reference corpus combines Gigaword 3 \cite{graff2007english} with a Wikipedia 2018 dump, lower-cased, yielding 5.4 billion tokens. We limit $\vocab_C$ and $\vocab_T$ to be the 50,000 most frequent tokens in $\corpus$.
We use a $5$-token context window, and $d=300$. We use the released implementations and hyperparameter choices of SGNS and GloVe. 
% (with number of negative samples $k\seq15$, undersampling probability of $2\sti10^{-5}$, unigram distribution smoothing of $0.75$) and GloVe (with empirical weighting function having $\alpha\seq\frac{3}{4}$ and $X_{max}\seq100$) as released in their implementations. 
Our implementation of \hilby uses PyTorch to take advantage of GPU-acceleration, automatic differentiation \cite{paszke2017automatic}, and the \textit{Adam} gradient descent optimizer \cite{kingma2014adam}. 
Practically, \hilby was implemented by using \textit{sharding} \cite{shazeer2016swivel}. We use a single 12-GB GPU, and load $12500 \times 12500$-element shards to calculate each update.  Training embeddings took less than 3 hours for a 50,000 word vocabulary. 
%For each model, we used an automatic learning-rate tuning algorithm that finds which learning rate facilitates rapid training while avoiding divergence. 
%We found the MF learning rates $\eta$ for each implemented algorithm to be: \hilby, $\eta\seq8.75$; SGNS-MF, $\eta\seq0.005$; GloVe-MF, $\eta\seq0.05$. 
\begin{table*}[t]
    \centering
    \begin{tabular}{l rrrr rrrr r}
    \toprule
        \textbf{Intrinsic eval.} & B143 & MENd/t & RMT & RARE & SE17 & S999 & WS-R/-S & Y130 & Analogy\\
    \midrule
        SGNS     & \textbf{.453} & .753/.763 & .680 & .515 & .656 & .364 & .589/.760 & .514 & \textbf{.763}/.312 \\
        %
        % SGNS-MF  & \textbf{.453} & .730/.754 & .650 & \underline{.559} & \textbf{.684} & \underline{.438} & .565/.751 & .498 & \textbf{.775}/\textbf{.354}\\
        %
        GloVe    & .347 & \textbf{.760}/\textbf{.771} & .663 & .509 & .662 & .391 & \textbf{.602}/.727 & \textbf{.541} & .739/.310\\
        %
        % GloVe-MF & .379 & \textbf{.764}/\underline{.768} & .664 & .526 & .671 & .419 & \textbf{.626}/.755 & \textbf{.574} & .745/.324 \\
        %
        \hilby   & .397 & .751/.761 & \textbf{.684} & \textbf{.579} & \textbf{.680} & \textbf{.462} & .593/\textbf{.765} & .514 & .705/\textbf{.343}\\
    \bottomrule
    \end{tabular}
    \caption{Performance on intrinsic evaluation datasets; first 8 columns are on word similarity tasks (\S\ref{sec:wordsim}), final column is analogy tasks (Google analogies/BATS) (\S\ref{sec:analogy}). The best result is in \textbf{bold}.}
    \label{tab:wordsim}
\end{table*}

% \begin{table*}[t]
%     \centering
%     \begin{tabular}{l rrrr rrrr}
%     \toprule
%         \textbf{Word sim.} & B143 & MENd/t & RMT & RARE & SE17 & SL999 & WS-R/-S & Y130\\
%     \midrule
%         SGNS     & \underline{.453} & .753/.763 & \underline{.680} & .515 & .656 & .364 & .589/\underline{.760} & .514 \\
%         %
%         MF-SGNS  & \textbf{.453} & .730/.754 & .650 & \underline{.559} & \textbf{.684} & \underline{.438} & .565/.751 & .498 \\
%         %
%         GloVe    & .347 & \underline{.760}/\textbf{.771} & .663 & .509 & .662 & .391 & \underline{.602}/.727 & \underline{.541} \\
%         %
%         MF-GloVe & .379 & \textbf{.764}/\underline{.768} & .664 & .526 & .671 & .419 & \textbf{.626}/.755 & \textbf{.574} \\
%         %
%         \hilby   & .397 & .751/.761 & \textbf{.684} & \textbf{.579} & \underline{.680} & \textbf{.462} & .593/\textbf{.765} & .514 \\
%     \bottomrule
%     \end{tabular}
%     \caption{Performance on word similarity datasets. Best is in \textbf{bold}, second best is \underline{underlined}..}
%     \label{tab:wordsim}
% \end{table*}

\subsection{Word similarity} \label{sec:wordsim}

A word similarity task involves interpreting the cosine similarity between embeddings as a measure of similarity or relatedness between words. Performance is computed with the Spearman rank correlation coefficient between the model's scoring of all pairs of words versus the gold standard human scoring. These tasks can reflect the degree of linear structure captured in the embeddings, which can provide useful insights into differences between models. However, they do not always correlate with performance in downstream tasks \cite{chiu2016intrinsic,faruqui2016problems}. 

We used the following word similarity datasets: 
Simlex999 (S999) \cite{hill2015simlex}; 
Wordsim353 \cite{finkelstein2002placing} divided into similarity (WS-S) and relatedness (WS-R) \cite{agirre2009study}; 
the SemEval 2017 task (SE17) \cite{camacho2017semeval}; 
Radinsky Mechanical Turk (RMT) \cite{radinsky2011word}; 
Baker Verbs 143 (B143) \cite{baker2014unsupervised}; 
Yang Powers Verbs 130 (Y130) \cite{yang2006verb};
MEN divided into a 2000-sample development set (MENd) and 1000-sample test set (MENt) \cite{bruni2012distributional}; 
Rare Words (RARE) \cite{luong2013better}. We had an average of 96\% coverage over all word-pairs in each dataset, excepting RARE; we had 31\% coverage over RARE, yielding 620 word pairs (i.e., more samples than SE17, RMT, WS-S/-R, and Y130). Results are computed on these covered word-pairs.

\paragraph{Word similarity results.}
Table~\ref{tab:wordsim} presents results across the 10 word similarity tasks. We observe that \hilby obtains the best performance in 5 out of 10 tasks. In particular, \hilby obtains substantially better scores on S999 than the other models, earning a Spearman correlation coefficient of 0.462, an $18\%$ relative improvement over the next best (SGNS). Note that S999 has been shown to have a high correlation with performance in extrinsic tasks such as Named Entity Recognition and  NP-chunking, unlike the other word similarity datasets \cite{chiu2016intrinsic}. On the tasks with worse performance, we observe that the differences between the three algorithms are relatively marginal. Base on these experiments, and the ones that follow, we can therefore conclude that our hypothesis (\S\ref{sec:experiments}) is valid.

\subsection{Analogical reasoning} \label{sec:analogy}
We performed intrinsic evaluation of our embeddings using standard analogy tasks (e.g., ``man'' is to ``woman'' as ``king'' is to $X$). We evaluated on the Google Analogy dataset (Google) \cite{mikolov2013efficient} and the Balanced Analogy Test Set (BATS) \cite{gladkova2016analogy}. We observed $86\%$ and $69\%$ coverage of the words in each dataset, respectively.
%, yielding 84,250 analogies total.
Preliminary experiments using \textit{3CosAdd} and \textit{3CosMul} \cite{levy2014linguistic} as selection rules, showed \textit{3CosMul} was always superior, consistent with the findings of \citeauthor{levy2014linguistic}.

% TODO (kian): reword this paragraph. DONE
\paragraph{Analogy results.} Table~\ref{tab:wordsim} presents results on the two analogy datasets in the final column. \hilby performs somewhat worse than the other models on the Google Analogy dataset. However, there has been a considerable amount of work finding that performance on these tasks does not necessarily provide a reliable judgment for embedding quality \citep{faruqui2016problems,linzen2016issues,rogers2017too}. Indeed, we can see that  performance on the Google Analogy dataset does not correspond with performance on the other larger analogy dataset (BATS), where \hilby gets the best performance.

\begin{table}[t]
    \centering
    \begin{tabular}{l r r}
        \toprule
        \textbf{Classification} & IMDB & AGNews \\
        \midrule
        SGNS     & .910 $\pm$ .001 & \textbf{.812} $\pm$ .003 \\
        % SGNS-MF  & \textbf{.912} $\pm$ .002 & \textbf{.813} $\pm$ .003 \\
        GloVe    & .905 $\pm$ .001 & .807 $\pm$ .003 \\
        % GloVe-MF & .904 $\pm$ .001 & .806 $\pm$ .002 \\
        \hilby   & \textbf{.911} $\pm$ .002 & \textbf{.812}  $\pm$ .003 \\
        \bottomrule
    \end{tabular}
    \caption{Classification results when using a BiLSTM-\textit{max} encoder. Best is \textbf{bold}.}
    \label{tab:classification}
\end{table}
\begin{table}[t]
    \centering
    \begin{tabular}{lccc}
    \toprule
        \textbf{Seq. labelling} & {Semcor} & {WSJ} & {Brown}\\
    \midrule
        \textit{Baseline} & \textit{.6126} & \textit{.8905} & \textit{.9349} \\
        \hline
        SGNS     & .6638  & .9615 & .9762  \\
        % SGNS-MF  & \textbf{.6696}  & \textbf{.9621} & \textbf{.9771} \\
        GloVe    & .6550  & .9609 & .9750 \\
        % GloVe-MF & .6655  & .9613 & .9759 \\
        \hilby   & \textbf{.6663} & \textbf{.9617} & \textbf{.9767} \\
    \bottomrule
    \end{tabular}
    \caption{Sequence labelling results when using a BiLSTM. Best is \textbf{bold}.}
    \label{tab:sequencelabelling}
\end{table}

% \begin{table}[t]
%     \centering
%     \begin{tabular}{lccc}
%     \toprule
%         \textbf{Seq. labelling} & {Semcor SST} & {WSJ POS} & {Brown POS}\\
%         (metric) & (micro-F1) & (acc.) & (acc.)\\
%     \midrule
%         Baseline & .6126 & .8905 & .9349 \\
%         \hline
%         SGNS     & .6638 $\pm$ .0014 & .9615 $\pm$ .0004 & .9762 $\pm$ .0002 \\
%         MF-SGNS  & \textbf{.6696} $\pm$ .0027 & \textbf{.9621} $\pm$ .0005 & \textbf{.9771} $\pm$ .0002 \\
%         GLoVe    & .6550 $\pm$ .0033 & .9609 $\pm$ .0004 & .9750 $\pm$ .0003 \\
%         MF-GLoVe & .6655 $\pm$ .0026 & .9613 $\pm$ .0003 & .9759 $\pm$ .0003 \\
%         \hilby   & \underline{.6663} $\pm$ .0028 & \underline{.9617} $\pm$ .0004 & \underline{.9767} $\pm$ .0002 \\
%     \bottomrule
%     \end{tabular}
%     \caption{Sequence labelling test set performance using a BiLSTM. We present mean accuracy plus/minus standard deviation across 10 runs with different random seeds for initialization. Best is in \textbf{bold}, second best is \underline{underlined}.}
%     \label{tab:sequencelabelling}
% \end{table}

\subsection{Text classification} \label{sec:classification}
We performed extrinsic evaluation for classification tasks on two benchmark NLP classification datasets. First, the IMDB movie reviews dataset for sentiment analysis \cite{maas2011learning}, divided into train and test sets of 25,000 samples each. Second, the AGNews
%%%  \footnote{\url{https://www.di.unipi.it/~gulli/AG_corpus_of_news_articles.html}} 
news classification dataset, as divided into 8 approximately 12,000-sample classes (such as \textit{Sports}, \textit{Health}, and \textit{Business}) by \citet{kenyon2018clustering}; here, we separate 30\% of the samples as the final test set. On each, we separate 10\% of the training set for validation tuning.
 
 We use a standard BiLSTM-\textit{max} sequence encoder for these tasks \cite{conneau2017supervised}. This model produces a sequence representation by max-pooling over the forward and backward hidden states produced a bidirectional LSTM. This representation is then passed through a MLP before final prediction. We found that validation performance was optimized with a 1-layer $128$-d BiLSTM, followed by a $512$-d MLP using a ReLU activation, a minibatch size of 64, dropout rate of $0.5$, and normalizing the embeddings before input. We use a learning rate of $0.001$ with Adam, and divide the learning rate by a factor of $10$ if validation performance does not improve in 3 epochs, similar to \citet{conneau2017supervised}; we schedule the learning rate in the same way for sequence labelling (\S\ref{sec:seqlab}).

\paragraph{Classification results.} In Table~\ref{tab:classification} we present the test set results from our classification experiments. We trained each BiLSTM-\textit{max} 10 times with different random seeds for weight initialization and present the mean test accuracy plus/minus the standard deviation.
These results show that each embedding model is similar, although GloVe is slightly worse than the others. Meanwhile, SGNS and \hilby perform approximately the same, obtaining high quality results on both tasks.

\subsection{Sequence labelling} \label{sec:seqlab}
Our final extrinsic evaluations are sequence labelling tasks on three datasets. The first task is \textit{supersense tagging} (SST) \cite{ciaramita2006broad} on the Semcor 3.0 corpus \cite{miller1993semantic}. SST is a coarse-grained semantic sequence labelling problem with 83 unique labels; we report results using the micro-F-score without the \textit{O}-tag score due to the skew of label distribution, as is standard \citep{alonso2017multitask,changpinyo2018multi}. We divide Semcor into a 70-30\% train-test split, and use 10\% of the training set for validation tuning. 
The second task is syntactic \textit{part-of-speech tagging} (POS); we use the Penn TreeBank Wall Street Journal corpus (WSJ) \cite{marcus1993building} and the Brown corpus (as distributed by NLTK\footnote{\url{https://www.nltk.org/book/ch02.html}}). For the WSJ, we use the given 44-tag tagset, and for Brown we map the original tags to the 12-tag ``universal tagset'' \cite{petrov2012universal}.
We use sections 22, 23, and 24 of the WSJ corpus as its test set, and separate out 30\% of the sentences in Brown as its test set.

On each dataset, we train a standard sequence labelling model inspired by \citet{huang2015bidirectional}: a 2-layer, $128$-d bidirectional LSTM, using a minibatch size of 16, and a dropout rate of 0.5. Interestingly, we found that normalizing the embeddings substantially reduced validation performance, so we keep them in their original form. 

\paragraph{Sequence labelling results.}
To accompany our results in Table~\ref{tab:sequencelabelling}, we include results from a trivial \textit{most-frequent-tag} baseline. This baseline returns that the tag of a token is the most frequently occurring tag for that token within the training set. In SST it is standard to include results from a most-frequent-\textit{supersense} baseline, being inspired from the tradition of word sense disambiguation, which uses the most-frequent-\textit{sense} baseline.

The results for the BiLSTMs are the mean test set score across 10 different runs with different random seeds for weight initialization. The low standard deviations were approximately the same for each model.
%; on Semcor, it was near .0025, on WSJ it was around .0003, and on Brown it was around .0002.
%
As in the classification tasks, we find that the embeddings produced by each model obtain very similar results. Nonetheless, we observe that \hilby offers marginal improvements over the others. Note that our performance on WSJ and Brown is expected since we use vanilla BiLSTMs that do not include any hand engineered character- or context-based features. Indeed, \citeauthor{huang2015bidirectional} report results of 96.04\% on the WSJ with their vanilla BiLSTM, which suggests that our embeddings possess strong syntactic properties.

\subsection{Qualitative analysis} \label{sec:vecscovecs}
% \textbf{T-T:} $\argmax_j \hvec{\textit{cat}} \cdot \hvec{j} =$  kittens, cats, kitten, poodle, terrier, dog \\
% \textbf{T-C:} $\argmax_j \hvec{\textit{cat}} j \rangle =$ burglar, siamese, scan, scans, tabasco, schr\"odinger\\
% \textbf{T-T:} $\argmax_j \hvec{\textit{money}} \cdot \hvec{j} =$ funds, monies, billions, cash, sums \\
% \textbf{T-C:} $\argmax_j \hvec{\textit{money}} j \rangle =$ laundering, launder, extort, laundered, funneling \\
% \textbf{T-T:} $\argmax_j \hvec{\textit{vector}} \cdot \hvec{j} =$ vectors, tensor, scalar, formula\_34, formula\_10 \\
% \textbf{T-C:} $\argmax_j \hvec{\textit{vector}} j \rangle =$ tangent, scalar, subspace, dimensional, non-zero.

\begin{table}[t]
    \centering
    \begin{tabular}{l l}
    \toprule
        \textbf{Argmax}$_i$ & {\textit{Top most similar embeddings}} \\
    \midrule
        $\hvec{i} \cdot \hvec{\textit{cat}}$ & kittens, cats, kitten, poodle \\
        % $\cos(\hvec{i}, \hvec{\textit{cat}})$ & kittens & cats & kitten & poodle \\
        $\hdot{i}{\textit{cat}}$ & burglar, siamese, schr{\"o}dinger \\
        $\hvec{i} \cdot \hvec{\textit{money}}$ & funds, monies, billions, cash \\
        $\hdot{i}{\textit{money}}$ & laundering, launder, extort \\
        $\hvec{i} \cdot \hvec{\textit{cuba}}$ & cubans, cuban, anti-castro \\
        $\hdot{i}{\textit{cuba}}$ & gooding, guantanamo, havana  \\
    \bottomrule
    \end{tabular}
    \caption{Qualitative analysis of the difference between the embedding recovery between vectors and vectors ($\hvec{i} \cdot \hvec{j}$) versus between vectors and covectors ($\hdot{i}{j}$).}
    \label{tab:vecscovecs}
\end{table}
We provide a final set of qualitative results in Table~\ref{tab:vecscovecs}. Here, we use the vectors and covectors trained by the \hilby model used in our experiments. These results elucidate the difference between using vector-vector similarity versus vector-covector dot product similarity (results are practically the same when using cosine similarity). The vector-vector similarity is well known as a way to measure semantic similarity between two concepts captured in word embeddings. As expected, we see recoveries like ``cat'' is similar to ``kitten'', ``money'' with ``funds'', etc. 

However, when we instead obtain the most similar \textit{covector} to the corresponding vector, the results are dramatically different.  We see that the vector for ``cat'' is most similar to covectors for words with which it forms multi-word expressions: ``cat burglar'', ''siamese cat'', ''schr{\"o}dinger's cat''.  We see that ``cuba'' is most similar to the covector for ``gooding'' -- this is because Cuba Gooding Jr. is a famous American actor whose Wikipedia page appears in our corpus. Indeed, the vector-covector recoveries are directly correlated to the PMIs between the terms in the corpus.

Overall, we see that vector-vector dot products recover semantic similarity, while vector-covector dot products recover co-occurrence similarity. \citet{melamud2015simple} and \citet{asr2017artificial} discuss these different statistical recoveries as \textit{paradigmatic} (target-to-target) and \textit{syntagmatic} (target-to-context) recoveries, respectively.  However, to our knowledge, previous work has not explicitly explained the reason for these two different types of recoveries; i.e., because the learning objective for word embeddings is to approximate PMI.  Therefore, these results qualitatively demonstrate exactly what our hypothesis anticipates: the dot product between trained vectors and covectors approximates the PMI between their corresponding words in the original corpus. 

\section{Discussion}
In the past, probabilistic distributional semantic models for creating word embeddings surpassed the traditional count-based models \cite{turney2010frequency} that preceded them, which was well-established by \citet{baroni2014don}. At the same time, models like Word2vec (SGNS), GloVe, and SVD of PPMI \cite{mikolov2013efficient,mikolov2013distributed,pennington2014glove,levy2015improving} offered strong improvements (in terms of performance and efficiency) over other probabilistically-motivated embedding models \cite{collobert2008unified,mnih2009scalable,turian2010word,mnih2013learning}. 

Today, NLP seems to be orienting toward deep contextualized models \cite{peters2018deep,devlin2018bert}. Nonetheless, pretrained word embeddings are still highly relevant. Indeed, they have been used recently to greatly assist solving problems in materials science \cite{tshitoyan2019unsupervised}, biomedical text mining \cite{zhang2019biowordvec}, and law \cite{chalkidis2019deep}.
Moreover, word embeddings are used in dynamic meta-embeddings to obtain state-of-the-art results \cite{kiela2018dynamic}, are used as inputs to ELMO \cite{peters2018deep}, and are crucial in memory-constrained NLP contexts (such as in mobile devices, which cannot store large deep neural networks \cite{shu2017compressing}). %\cite{yogatama2015learning,faruqui2015sparse}

We believe a robust understanding of the ``shallow'' (or, non-deep), uncontextualized embedding models presented in this work is a prerequisite for informed development of deeper models.
In this work, we advanced the theoretical understanding of word embeddings by proposing the \textit{low rank embedder} framework. 
Cast under this framework, the similarities between many existing algorithms become apparent.  
After isolating two key principles shared by the low rank embedders---a probabilistically informed bilinear parameterization of PMI and a ``tempered'' gradient conditioning measure---we demonstrate that these ideas are sufficient to derive \hilby, a model that is simpler yet has performance that is equivalent or better than the other models. 
This provides a parsimonious explanation for the success of the low rank embedders, demonstrating that many idiosyncratic features of other embedders are unnecessary.  

The design parameters of our framework have not yet been fully explored.  Our framework could be used to model more complex linguistic phenomena, by following the blueprint provided by our derivation of \hilby.  Moreover, based on our findings concerning the importance of covectors in parameterizing the model's approximation of PMI, we believe that methods such as retrofitting \cite{faruqui2015retrofitting},  subword-based embedding decomposition \cite{stratos2017reconstruction}, and dynamic meta-embedding \cite{kiela2018dynamic} could benefit by incorporating covectors into their modelling designs.  As well, similar to how the theoretical basis of LDS \cite{arora2016latent} was the grounding for the widely-used SIF document embedding algorithm \cite{arora2017simple}, we believe that the theoretical basis provided in this work can inform future development of document embedding techniques.

\bibliography{tacl2018}
\bibliographystyle{acl_natbib}

\end{document}